\documentclass{ecai} 

\usepackage{latexsym}
\usepackage{amssymb}
\usepackage{amsmath}
\usepackage{amsthm}
\usepackage{booktabs}
\usepackage{enumitem}
\usepackage{graphicx}
\usepackage{color}
\usepackage{algorithm}
\usepackage{algorithmic}

\usepackage{multirow}
\newtheorem{theorem}{Theorem}

\newtheorem{proposition}[theorem]{Proposition}

\newcommand{\BibTeX}{B\kern-.05em{\sc i\kern-.025em b}\kern-.08em\TeX}

\begin{document}


\begin{frontmatter}


\paperid{9787} 


\title{Robustness Feature Adapter for \\
Efficient Adversarial Training}




\author[A]{\fnms{Quanwei}~\snm{Wu}}
\author[A]{\fnms{Jun}~\snm{Guo}}
\author[B]{\fnms{Wei}~\snm{Wang}}
\author[A]{\fnms{Yi}~\snm{Wang}\thanks{Corresponding Author. Email: wangyi@dgut.edu.cn}}
\address[A]{Dongguan University of Technology}
\address[B]{The Hong Kong University of Science and Technology (Guangzhou)}

\begin{abstract}
Adversarial training (AT) with projected gradient descent is the most popular method to improve model robustness under adversarial attacks. However, computation overheads become prohibitively large when AT is applied to large backbone models. AT is also known to have the issue of robust overfitting. This paper contributes to solving these problems coherently towards building more trustworthy foundation models for multimedia applications. In particular, we propose a new adapter-based approach for efficient AT directly in the feature space. We show that the proposed adapter-based approach can improve the inner-loop convergence quality by eliminating robust overfitting. As a result, it significantly increases computational efficiency and improves model accuracy by generalizing adversarial robustness to unseen attacks. We demonstrate the effectiveness of the new adapter-based approach in different backbone architectures and in AT at scale.

\end{abstract}

\end{frontmatter}

\section{Introduction}
\label{sec:intro}

Deep neural networks (DNNs) are popular foundation models in many multimedia applications. Despite their prevalence, even large vision and language models (VLMs) are vulnerable to adversarial example attacks \cite{Zhou2023a},
where visually imperceptible adversarial perturbations are added, typically to the input examples, to induce model malfunctions.  

Adversarial training (AT) is by far the most effective defense against such attacks \cite{croce2020reliable}. It is formulated as a Min-Max problem \cite{Madry2018} where the inner maximization is to find the ``worst cases'' to attack and the outer minimization updates the model parameters to account for the findings. 
As computing the optimal examples is NP hard \cite{andriushchenko_understanding_2020}, the worst cases can only be approximated, for example by using the PGD attack \cite{Madry2018}. 
In fact, AT-PGD and its variants demonstrate the best empirical robustness of AT in practice \cite{croce2020reliable}. 
In the Min-Max framework, AT generates adversarial examples (AEs) in the run and adds them back to each training epoch for iterative optimization. 
Recent studies show that AT requires substantially more training data to gain generalization. Thus, training time can become prohibitively slow to hinder the use of AT by scaling up for VLMs \cite{Vatsa2024}. 

There were efficient AT methods proposed in the literature by recycling the use of gradient information \cite{shafahi2019adversarial}, single-step 
AT with FGSM \cite{wong2020fast}, and partial AT such as early stop of PGD \cite{zhang_attacks_2020} and generating AEs on more effective subsets of data \cite{Hua2021}. 
However, these methods often encounter \emph{overfitting} issues that cause poor robust generalization to unseen attacks \cite{Rice2020}.

Specifically, robust overfitting (RO) occurs in AT after a certain point where the training robustness keeps increasing but the test robustness decreases. 
Unlike in standard training (ST), the RO problem cannot be easily solved by simply increasing the size of the training data.  
In \cite{Wang2023overfit}, it is conjectured that RO is the effect of the local fitting ability that overpowers the attack ability. The authors further explain the phenomenon from the prevailing understanding of AT by the composition of robust and non-robust features that are both useful for classification \cite{Ilyas2019bugs}. As a result, the overfitting may be caused by false memorization of non-robust features in the later stage of AT when the model becomes more robust.

Inspired by this insight, we propose in this paper to retain non-robust features in AT by separating the learning paths of the two players. 
This is done by introducing an external module, the so-called robustness feature adapter (RFA), into a backbone model such that the inner loop attacker can generate stronger attacks while the outer loop trainer performs a parameter-efficient fine-tuning (PEFT) \cite{Houlsby2019} version of AT that distills more robust features for the modeling task with RFA.   
Note that adapter-based methods for fine-tuning pre-trained models are emerging for downstream tasks \cite{zhang2024adapt}.

To the best of our knowledge, this is the first work in which PEFT is adapted to facilitate AT in feature space.
For the threat model, we consider gray-box attacks generated on pre-trained backbone models. The attack setting is not uncommon as many pre-trained models are publicly available for developing multimedia downstream applications \cite{Vatsa2024}. 
Thus, attackers can get easy access to the pre-trained model and transfer attacks to downstream tasks without knowing the fine-tuning procedures.
The low complexity network also makes it feasible to secure the external adapter by DNN encryption schemes \cite{Lee2022}.
Even in the worst-case scenario where the series adapter is compromised, we show that the adversarial robustness is
simply reduced to that of the backbone model, whether or not being robustified.

The main contributions of this paper are:
\begin{itemize}

\item 
We present an efficient and flexible solution to robustify an open large model by incorporating an external adapter so-called RFA to fine tune the backbone model. 

\item 
We propose to improve robust generalization of AT by generating more robust AEs directly in the feature space, and provide a theoretical analysis on how latent perturbation can affect the inner-loop maximization process.

\item 
We advocate the critical role of non-robust features in robust generalization of AT and, accordingly, design the RFA training with two driven classifiers and a triplet loss for the outer-loop minimization process.

\end{itemize}

\section{Related Work}
\label{sec:RW}

The conventional threat models consider \emph{white-box} attacks that generate AEs by having full access to the entire model \cite{Goodfellow2015}. In real-world applications, however, some aspects of a defense may be held secret to the defender \cite{Athalye2018}. Attacks generated in such cases are called \emph{gray-box} or quasi-black-box settings \cite{Athalye2018,Yang2021b,Zhou2023a}. With more pre-trained models unveiled to the public, there is an increasing risk of downstream-agnostic attacks in the gray-box setting where the attacks are generated on pre-trained models and  transferred to attack the downstream tasks \cite{Zhou2023a}.

With pre-trained models, downstream defenses can be categorized into three types of strategies, namely pruning or adversarial distillation, data pre-processing by corruption, 
and AT with downstream tasks. 
In particular, adversarial distillation aims to \emph{inherit} model accuracy and adversarial robustness from a robust pre-trained model, typically with a teacher-student framework  \cite{Huang2023}.
Corruption-based methods aim to purify or corrupt adversarial perturbations in the pre-processing phase, usually at the model input \cite{zhang2024classifier,Yang2021b}. 
They can be processed in gray box settings and incorporated with other defense strategies such as AT to achieve SOTA performance \cite{Yang2021b,zhou2023phase}.

AT with fine-tuning procedures is most effective by itself \cite{Zhou2023a}. However, AT has inherent problems with computational overheads and robust generalization \cite{Bai2021}. Both can become prohibitive when the complexity of the network increases \cite{Vatsa2024}.
\cite{pang_robustness_2022} attributes the problems to an improper definition of robustness. The authors proposed a method called SCORE to reconcile robustness and accuracy, achieving \emph{top-rank} performance at the time.
With the recent breakthrough of generative models such as DDPM \cite{ho_denoising_2020}, learning-based data augmentation has been incorporated into AT \cite{gowal_improving_2021,wang_better_2023}. There are also investigations to impose regularization on generating adversarial data to improve the generalizability of the model \cite{pmlr-v202-yang23h}. 

Existing methods that aim to reduce model vulnerability by learning more robust features \cite{yang_disentangle_2021,Li2023} tend to overlook the RO problem by either ignoring or converting the non-robust components, and do not generalize well to unseen attacks. 
Note that non-robust features are not only fragile but also useful for model generalization \cite{Ilyas2019bugs}. 
Recently, theoretical progress was made on understanding RO \cite{Wang2023overfit}. By viewing AT as a dynamic game, it conjectures that RO is a result of false memorization of non-robust features in the inner-loop optimization process. The falsely mapped non-robust features in training data do not generalize to improve the test robustness of AT.
To resolve the problems, the authors proposed to re-balance the two min-max players by either regularizing the trainer's capacity or improving the attacker's strength.

\begin{figure*}[t!]
	\centering
        \includegraphics[width=.86\textwidth]{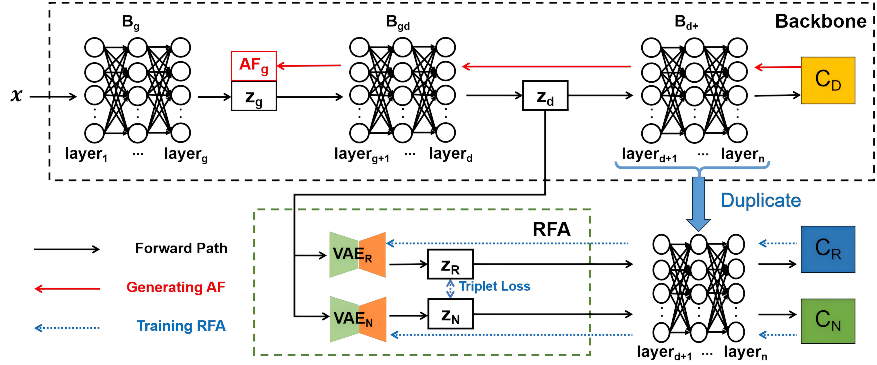}
	\caption{The proposed feature-based AT framework with RFA.}
	\label{fig:RFA-train}
\end{figure*}

\section{Methodology} 

We first briefly review the general min-max framework of AT where the inner-loop attacker tries to maximize the empirical risk $\mathcal{L}$ while the outer-loop trainer updates model parameters $\theta$ to account for the changes over all in-distribution $x\in\mathcal{D}$. It can be formulated as
\begin{equation}
	\min_{\theta} \mathbb{E}_{(x, y)\sim\mathcal{D}}\left[\max_{\|\delta\|<\epsilon} 	\mathcal{L}(\theta,x+\delta,y)\right]
	\label{eq:min-max}
\end{equation} 
where the perturbation $\delta$ is added within a certain budget $\epsilon$, and $\mathcal{L}(\cdot)$ is typically the cross-entropy loss.
It starts to incur RO at a later stage of AT when the AEs in the test become out-of-distribution (OOD) from those generated by the inner attacks. 
To resolve this problem, we propose a simple solution by incorporating RFA to the backbone as shown in Figure \ref{fig:RFA-train}.
The following sections specify the proposed approach for the inner and outer optimization of AT, respectively.

\subsection{Inner-Loop Maximization}
\label{sec:inner}

The inner-loop maximization is a constrained optimization problem, and is in general globally non-concave. 
The goal is to find ``worst cases'' by introducing perturbations that increase the empirical risk $\mathcal{L}$ for outer optimization. 
Unlike conventional methods, we propose to impose $\delta$ directly in the feature space instead of at the input $x$. We shall show later that $\delta$ is amplified through the network layers in the forward propagation path to cause erroneous outputs.

Denote the backbone by $B$ and the intermediate feature of $x$ by $z_g=B_g[x]$ which is the output of the $g$-th network layer. 
Similarly, $z_d=B_d[x]=B_{gd}[z_g]$ indicates those at the $d$-th layer where $B_{gd}$ is the network fraction between the $(g+1)$-th and the $d$-th layer.
As shown in Figure \ref{fig:RFA-train}, the subsequent network fraction after the $d$-th layer is denoted by $B_{d+}$ through which $z_d$ is propagated to the output layer to get the soft label $\hat{y}=B_{d+}[z_d]=B[x]$. The backbone model is then expressed as $B=B_{g}\circ B_{gd}\circ B_{d+}$. 

Without loss of generality, we define the optimal perturbation added to $z_g$ by 
\begin{equation}
\delta^* := \arg\max_{z'_g\in \mathcal{S}(z_g)} \mathcal{L}\left(B_{g+}[z'_g], y\right)
\label{eq:delta}
\end{equation}
where we call $z'_g = z_g+\delta_g$ as adversarial features (AFs) with $g>0$ on the backbone. 
To evaluate $\delta^*$, we adopt $k$-step PGD \cite{Madry2018} for approximation. Note that it is possible to use other numerical solutions for evaluation, too.
Accordingly, the feature perturbation is added iteratively by
\begin{equation}
	\delta^{(k)}_g = \alpha_g \cdot \texttt{sign}\left(\nabla_{z_{g}} \mathcal{L}(B_{g+}[z_{g}^{(k-1)}], y)\right)
	\label{eq:g}
\end{equation}
where $z_g^0:=z_g$ and $z_{g}^{(k)}\in \mathcal{S} (z^0_g)$ with $\mathcal{S}(z_g^0) = \{z'_g | \|z'_g - z_g^0\|_\infty < k\alpha_g \}$ and $\alpha_g=\eta_g\mu(|z_g|)$ is the step size. 
Similar to BN \cite{ioffe2015batch}, we restrict $\delta_g$ with respect to a proportion, controlled by the scaling factor $\eta_g$ of the statistical mean value $\mu(|z_g|)$ in each mini-batch to account for feature distribution at different layers. 
We find $\eta_g$ empirically for each layer, e.g., $\eta_1 = 0.01$ and $\eta_3=0.035$.  

We now study the effect of latent perturbation at different sub-layers by selecting $g$ for the inner optimization of AT. 
Denote the variation of empirical risk $\mathcal{L}$ by 
\begin{equation}
   \Delta\mathcal{L}_g = \max_{\|\delta_g\|_\infty \leq k\cdot\alpha} | \mathcal{L}(B_{g+}[z'_g], y) - \mathcal{L}(B_{g+}[z_g], y) | ~.
\end{equation}

\begin{figure}[t!]
	\centering
	\begin{minipage}{0.49\linewidth}
		\centering
		\includegraphics[width=\linewidth]{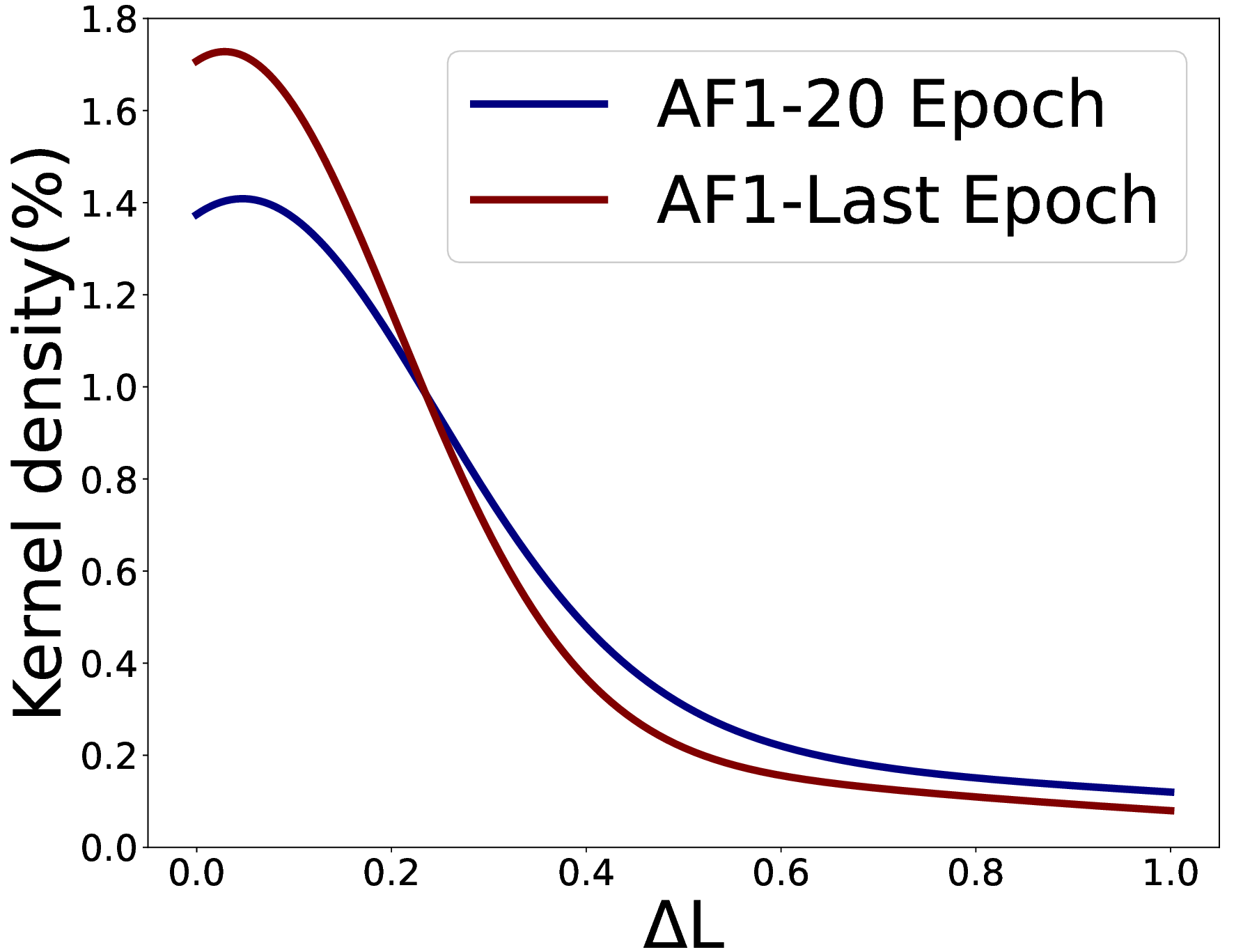} 
		\textbf{(a)} $g=1$
	\end{minipage}
	\hfill
	\begin{minipage}{0.49\linewidth}
		\centering
		\includegraphics[width=\linewidth]{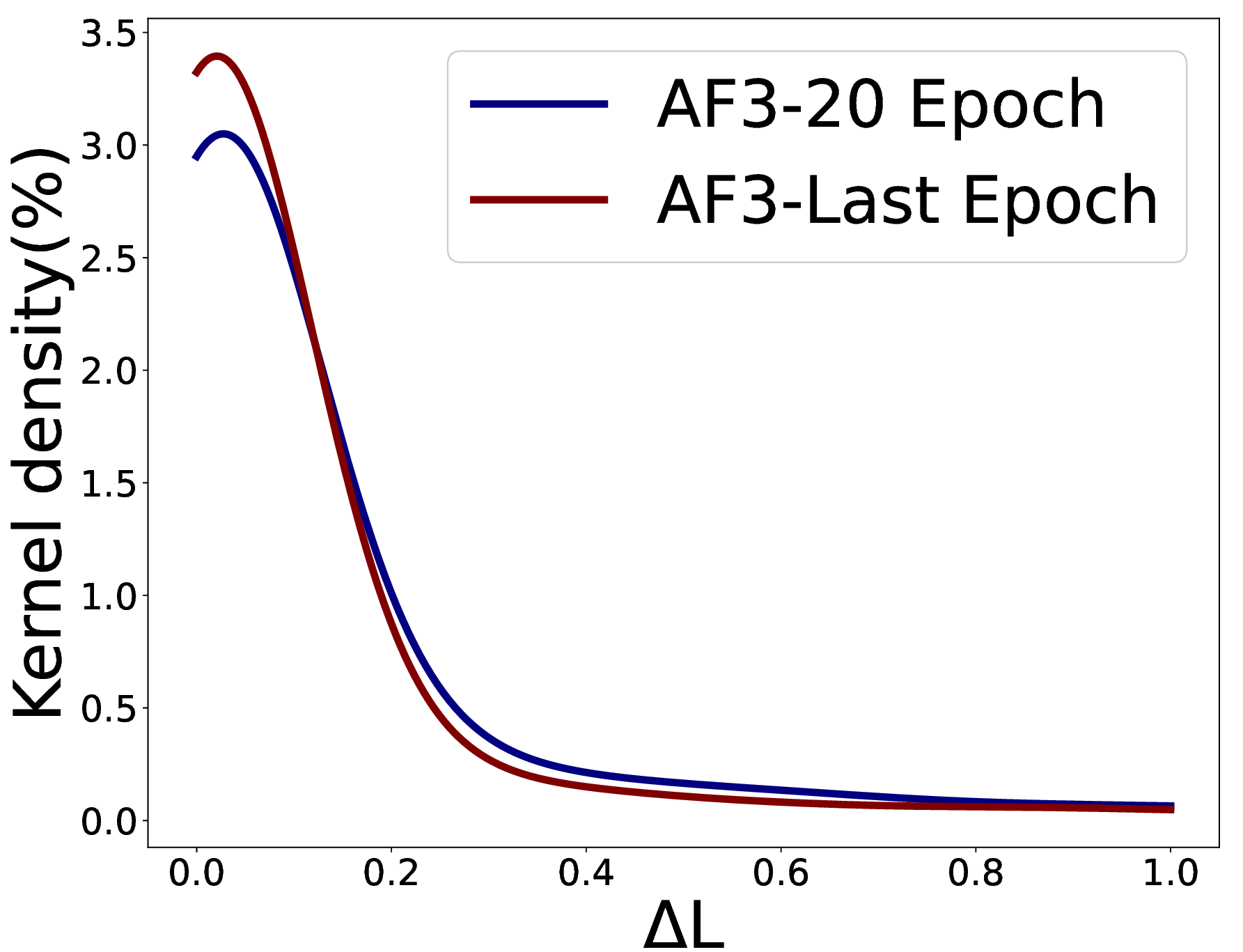} 
		\textbf{(b)} $g=3$
	\end{minipage}
	\caption{Distribution of $\Delta L_g$ due to $\delta_g$ in RFA training.}
	\label{fig:DeltaLk4Train}
\end{figure}

\begin{proposition}
Given the same $\|\delta_g\|_\infty<k\alpha$, perturbing lower level features, i.e., with a smaller $g>0$, tends to cause a larger $\Delta\mathcal{L}_g$.
\end{proposition}
In Appendix A1, we show that the above proposition holds under certain assumptions.  
Figure \ref{fig:DeltaLk4Train} verifies the proposition by computing the distribution of $\Delta\mathcal{L}_g$ with different $g$, using Gaussian kernel density estimation. In all cases, the step number and the scaling factor are set to the same with $k=10$ and $\eta = 0.01$. It can be seen that perturbing at $g=1$ tends to cause a larger $\Delta\mathcal{L}$ than at $g=3$, especially at an early stage of AT. This indicates a better convergence quality \cite{Wang2019FOSC}, hence a better trade-off of training efficiency and robustness.
Section \ref{sec:g} will discuss the effect of $g$ for inner optimization. 

\subsection{RFA Training}
\label{RFA}

It was observed that feature representations learned to correctly predict and robustly predict are different \cite{Ilyas2019bugs}. 
Thus, it is important to use robust features in the outer-loop minimization process. 
In addition, we propose to impose a constraint on the distilled robust features such that they are \emph{pushed away} from the estimated distribution of non-robust features that are prone to inner-loop attacks. 
The idea is implemented with a two-classifier-driven double-VAE network for RFA as shown in Figure \ref{fig:RFA-train}.

By perturbing $z_g$ at the $g$-th layer, the adversarial features 
$z'_g$ is added to the training data in each mini-batch. 
Suppose that RFA is embedded to the $d$-th layer, for $d>g$, so that it can work on data under perturbation. 
The impact of choosing different $d$ for the embedding is discussed in Section \ref{sec:d}. 
Let the normal features $z^{+}_d=B_{gd}[z_g]$ and the adversarial features $z^{-}_d=B_{gd}[z'_g]$. 
We adopt two simple VAE networks for RFA to learn the robust and non-robust components of $z^{+}_d$ and $z^{-}_d$, denoted by $z^{+}_\mathtt{R}, z^{-}_\mathtt{R}, z^{+}_\mathtt{N}, z^{-}_\mathtt{N}$, respectively.

For classification, robustness is often specified as whether the class-representation features are changed under adversarial perturbation. Thus, we deploy two classifiers, denoted by $\mathtt{C_R}$ and $\mathtt{C_N}$, to drive the RFA learning.
In particular, we duplicate $B_{d+}$ to build $\mathtt{C_R}$ and $\mathtt{C_N}$ by simulating the performance of classifier $\mathtt{C_D}$ on the backbone.
Specifically, 
we define the cross-entropy (CE) loss by 
\begin{eqnarray}
&\mathcal{L}_{\mathtt{C_R}} := \mathtt{CE}\left(\mathtt{C_R}(B_{d+}[z^{+}_\mathtt{R}]), y\right) + \mathtt{CE}(\mathtt{C_R}\left(B_{d+}[z^{-}_\mathtt{R}]), y\right)
\label{eq:ce1} \\
&\mathcal{L}_{\mathtt{C_N}} := \mathtt{CE}\left(\mathtt{C_N}(B_{d+}[z^{+}_\mathtt{N}]), y\right) + \mathtt{CE}\left(\mathtt{C_N}(B_{d+}[z^{-}_\mathtt{N}]), \bar{y}\right)
\label{eq:ce2}
\end{eqnarray}
where $y$ is the ground-truth label and $\bar{y}$ is the erroneous prediction label produced by $\mathtt{C_D}$ under untargeted attacks.

In addition, we introduce triplet loss to the RFA learning by increasing the distance between the robust and non-robust components of perturbed features. 
The triplet loss function is defined by \cite{khosla_supervised_2020} as
\begin{equation} 
    \mathtt{Tp}(a, p, n, \tau) = \max \{\|a-p\|^2 - \|a-n\|^2 + \tau, 0\}
    \label{eq:triplet_loss}
\end{equation}  
where $a$ is the anchor point, $(p, n)$ are positive and negative sample pairs, and $\tau$ is empirically defined for the margin between $p$ and $n$. 
We propose to use the robust components of normal features ($z^{+}_\mathtt{R}$) as the anchor points and the robust components of adversarial features ($z^{-}_\mathtt{R}$) as positive samples. Since there are two types of negative samples, i.e., $z^{+}_\mathtt{N}$ and $z^{-}_\mathtt{N}$, for non-robust features, we propose the following function for jointly training the RFA networks: 
\begin{equation} 
 	\mathcal{L}_{\mathtt{Tp}} := \mathtt{Tp}(z^{+}_\mathtt{R}, z^{-}_\mathtt{R}, z^{+}_\mathtt{N}, \tau) + \mathtt{Tp}(z^{+}_\mathtt{R}, z^{-}_\mathtt{R}, z^{-}_\mathtt{N}, \tau) ~.
 	\label{eq:tp}
 \end{equation} 
Note that both robust and non-robust features play critical roles in the RFA training, especially in \eqref{eq:tp} which aims to push $z^{-}_\mathtt{R}$ away from non-robust features that may be exploited to generate attacks.
The feature decoupling effects of RFA and their semantic changes under attacks can be visualized in Appendix A2.

\subsection{Outer-Loop Minimization}

Denote model parameters for the backbone and RFA by $\theta_B$ and $\theta_A$, respectively. 
We can perform the outer-loop minimization process in two modes: 1) RFA-FB by updating $\theta_A$ only, and 2) RFA-UB by updating both $\theta_A$ and $\theta_B$. 
 
\noindent\textbf{RFA-FB:} This mode trains $\theta_A$ only by fixing the backbone. It works in the same way as the parameter-efficient fine-tuning (PEFT) \cite{Houlsby2019} which become popular for efficiently fine-tuning large pre-trained models such as LLMs with a small set of external parameters while still achieving comparable or even superior performance. 
By inserting the adapter into the $d$-th sublayer of the backbone, the RFA training data $z^{*}_{d}$ consists of both normal and adversarial features with supervised information as described in Section \ref{RFA}. 
By fixing $\theta_B$, it is an incarnation of PEFT for AT by training RFA with 
\begin{equation}
 \mathcal{L}_\texttt{FB} (\theta_A) = \mathcal{L}_{\mathtt{C_R}} + \lambda_\mathtt{C_N}  \mathcal{L}_{\mathtt{C_N}} +  \lambda_\mathtt{Tp}  \mathcal{L}_{\mathtt{Tp}}
 \label{eq:FB}
\end{equation}
where the hyper-parameters balance the CE loss and the triplet loss defined in \eqref{eq:ce1}-\eqref{eq:tp}. The outer optimization by the trainer is thus to find $\theta^*_A := \arg\min  \mathcal{L}_\texttt{FB}$.

\noindent\textbf{RFA-UB:} This mode jointly updates $\theta:=\theta_A\cup\theta_B$ in a recursive manner. In particular, we first update $\theta_A$ using \eqref{eq:FB} by fixing $\theta_B$, and then update $\theta$ with  
\begin{equation}
 \mathcal{L}_\texttt{UB} (\theta) = \mathcal{L}_{\mathtt{C_R}}  +  \lambda_\mathtt{Tp}  \mathcal{L}_{\mathtt{Tp}}  + \lambda_{\mathtt{B}}  \mathcal{L}_{\mathtt{B}} 
 \label{eq:UB}
\end{equation}
where $\mathcal{L}_{\mathtt{B}} := \mathtt{CE}\left(\mathtt{C_D}(B[x]), y\right)$ is the CE loss by the backbone classifier.
To align the backbone learning more with $\mathtt{VAE_R}$ for robust features, 
we do not include  $\mathcal{L}_{\mathtt{C_N}}$ in \eqref{eq:UB} as the noisy labels $\bar{y}$ may cause interference to the backbone learning.
The joint optimization is performed with $\mathcal{L}_{\mathtt{FB}}$ and $\mathcal{L}_{\mathtt{UB}}$ alternately in a 1:1 fashion.

\subsection{Inference with RFA}
\begin{figure}[t!]
	\centering
        \includegraphics[width=\linewidth]{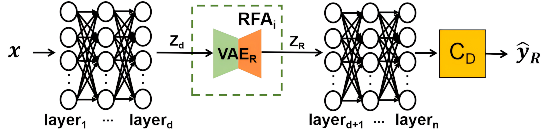}
	\caption{Inference with RFA.}
	\label{fig:RFA-inference}
\end{figure} 
Once trained with a backbone, the RFA module can be simplified to $\mathtt{VAE_R}$ by eliminating $\mathtt{VAE_N}$ and the classifier components. 
This is because only the distillation of robust features is needed for RFA inference with the backbone, namely RFA$_i$ shown in Figure \ref{fig:RFA-inference}.
This enables a flexible way for adversarial detection by 
comparing the difference of predictions before and after the RFA embedding. 
Moreover, the external adapter module can be encrypted to improve the model's security. This is feasible because RFA$_i$ is small with a low-complexity network, e.g., having only two fully connected or convolution layers for the encoder and decoder of $\mathtt{VAE_R}$. Thus, it is possible to secure it using DNN encryption schemes such as \cite{Lee2022}. This enables great flexibility and portability to imbue a large pre-trained model with adversarial robustness for downstream tasks.
\section{Discussions}
\label{Analysis}

\subsection{Effects of $g$ for Adding Perturbations}  
\label{sec:g}

Proposition 1 indicates that, with the same perturbation budget, perturbing deep features closer to the input results in stronger attacks in the min-max game, hence better model robustness by AT.
In this section, we perform verification tests to evaluate the effects of generating latent perturbations at different sub-layers.
The experiments are performed with RFA-FB on the backbone model of WRN-28-10 pre-trained on CIFAR-10. The results are shown in Table \ref{tab:varying_g}.

For reference, we also provide results by generating perturbation on $X$ at the input image level, i.e., $g=0$. Note that the perturbation size of AE is typically small, e.g.,  $\|\delta_0\|_\infty \leq \epsilon = 8/255$, in order to fool human eyes.
Whereas there is no mandatory requirement of imperceptibility for AF, which improves inner-loop convergence quality hence better generalizability of AT by perturbing features.

Denote the first-layer feature (i.e., $g=1$) by $z^{+}_1$ and $z^{-}_1$ for before and after adding $\delta_0$ to the input, respectively. Thus, the perturbation effect of $\delta_0$ on $z_1$ can be estimated by $\|\tilde{\delta}_1\|_1 = \|z^{-}_1 - z^{+}_1\|_1$.

To impose a similar effect on $z_1$ as that by $\delta_0$, we estimate the perturbation budget by directly perturbing at $g=1$, i.e., $\max \|\delta_1\|_\infty = k\alpha_1 $, in the following way. 
We first compute all $\|\tilde{\delta}_1\|_1$ as described above for the normal and adversarial example pairs over 500 mini-batches with a batch size of 100. 
Let 
$k\alpha_1 = k\eta\cdot\mu(|z_1|) = \|\tilde{\delta}_1\|_\infty =\max \|\tilde{\delta}_1\|_1$ with $k = 10$ for the PGD steps. 
We estimate $\mu(|z_1|)$ by the statistical mean of $|z^{+}_1|$ from which we can find an approximate value for $\eta$, e.g., $\eta_1 = 0.01$ for $g=1$ in this case.
Table \ref{tab:varying_g} shows the results by fixing $k\eta=1$ for feature perturbation at AF$_1$, AF$_2$, and AF$_3$, respectively, to compare the perturbation effect at different layers with a similar budget.

It can be seen that the test robustness tends to be higher with a smaller $g$. This is consistent with previous observations that AT with stronger attacks yields better adversarial robustness \cite{Bai2021,Wang2023overfit}. 
In terms of efficiency and complexity, however, the advantage is also obvious to perform inner optimization in the higher-level feature space by selecting a larger $g$ closer to the output layer.

\begin{table}[t!]
	\centering
	\caption{Test time and robustness of AT-RFA by selecting $g$. }
	\begin{tabular}{p{0.5cm}|c|c|p{0.8cm}|p{0.45cm}p{0.4cm}p{0.4cm}p{0.95cm}} 
		\bottomrule
		\multirow{2}{*}{AF$_g$} & \multirow{2}{*}{\begin{tabular}[c]{@{}c@{}}Time\\(s)\end{tabular}} & \multirow{2}{*}{\begin{tabular}[c]{@{}c@{}}Clean \\Acc.\end{tabular}} & \multicolumn{5}{c}{Robust Accuracy}  \\ 
		\cline{4-8}
		&                                                                    &                                                                       & PGD$_\infty$  & PGD$_2$  & AA$_\infty$   & AA$_2$  & ReColor  \\ 
		\hline
		AE      & 1.187  		 & 94.3 & \textbf{97.6}                           & 94.6 & \textbf{93.8} & \textbf{92.7} & \textbf{77.8} \\
		AF$_1$  & 1.184 		 & \textbf{94.4} & 97.1 & \textbf{95.1} & 92.9 & 90.3 & 74.1 \\
		AF$_2$  & 0.745 		 & 92.6 & 71.2 & 78.9 & 67.5 & 62.3 & 61.8 \\
		AF$_3$  & \textbf{0.444} & 94.1 & 39.5 & 58.9 & 37.8 & 32.0 & 29.2 \\
		\toprule
	\end{tabular}
	
	\label{tab:varying_g}
\end{table}

\begin{table}[t!]
	\centering	
	\caption{Decoupling effects and time cost by selecting $d$.}
	\begin{tabular}{p{1.1cm}|c|p{0.65cm}|p{0.1cm}p{0.65cm}p{0.5cm}p{0.5cm}p{0.7cm}} 
		\bottomrule
		\multirow{2}{*}{Mode} & \multirow{2}{*}{\begin{tabular}[c]{@{}c@{}}$d$\end{tabular}} & \multirow{2}{*}{\begin{tabular}[c]{@{}c@{}}Time\\(s)\end{tabular}} & \multicolumn{5}{c}{MIC($z_R$,$z_N$)}                                                                       \\ 
		\cline{4-8}
		&                                                                   &                                                                    & \multicolumn{1}{l}{Clean} & PGD$_\infty$            & AA$_2$            & CW$_2$            & FGSM            \\ 
		\hline
		\multirow{2}{*}{RFA-FB}   & 3                                                                 & 1.314                                                              & \textbf{0.032}           & \textbf{0.020} & \textbf{0.020} & \textbf{0.023} & \textbf{0.021}  \\
		& 4                                                                 & \textbf{1.187}                                                     & 0.506                    & 0.35           & 0.359          & 0.446          & 0.317           \\ 
		\hline
		\multirow{2}{*}{RFA-UB} & 3                                                                 & 2.271                                                              & \textbf{0.073}           & \textbf{0.053} & \textbf{0.052} & \textbf{0.054} & \textbf{0.050}  \\
		& 4                                                                 & \textbf{2.156}                                                     & 0.239                    & 0.125          & 0.126          & 0.133          & 0.115           \\
		\toprule
	\end{tabular}
	
	\label{tab:varying_d}
\end{table}

\subsection{Effects of $d$ for Embedding The RFA} 
\label{sec:d} 

The embedding position of RFA $d$ affects feature decoupling and semantic changes under attacks. The optimal value in our design is subject to the trade-off between the decoupling effects and computational efficiency.
In this section, we evaluate both operation modes of RFA-FB and RFA-UB with AE on WRN-28-10 over CIFAR-10.
The decoupling effects of feature disentanglement are evaluated with the quantitative criterion of Maximum Information Coefficient (MIC) \cite{albanese2018}, which ranges between 0 and 1, with a smaller value closer to 0 indicating a weaker correlation between feature vectors.
Table \ref{tab:varying_d} shows the quantitative results of decoupling effects between the robust and non-robust features $z_r$ and $z_n$. 
Both RFA-FB and RFA-UB achieve better decoupling results at $d=3$ when the embedding position of RFA gets closer to the input layer.
However, our experiments show that in practice, the RFA embedding at $d=4$ is able to achieve good enough robustness performance but significantly increases the computation efficiency of AT. We use $d=4$ for WRN-28-10 in the following experiments unless otherwise specified. 
The feature decoupling effects and their semantic changes under attacks can be visualized in Appendix A2.

\subsection{Robust Overfitting}

\begin{figure}[t!]
	\centering
	\begin{minipage}{0.49\linewidth}
		\centering
            \includegraphics[width=\linewidth]{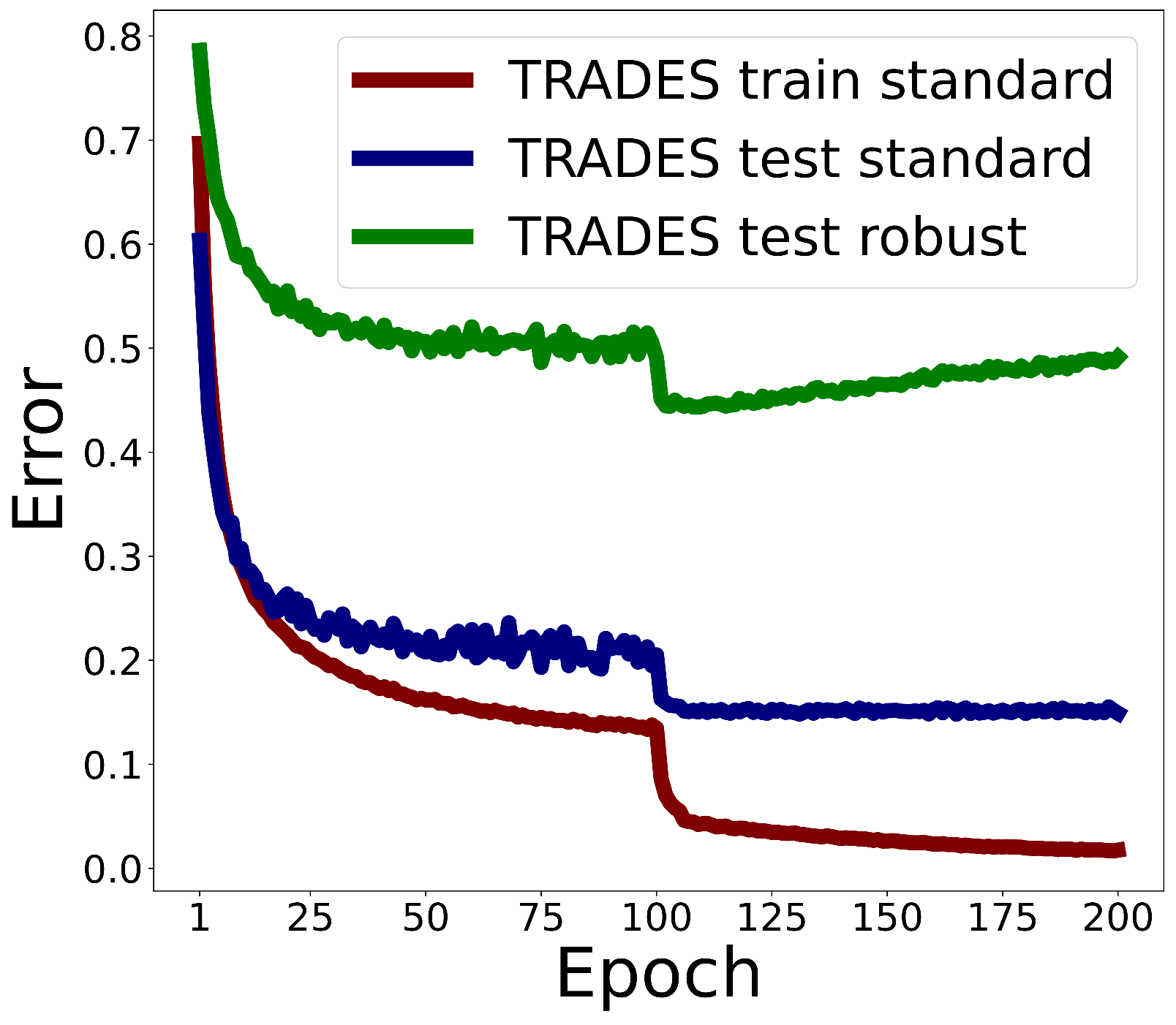}
		\textbf{(a)} TRADES
	\end{minipage}
	\hfill
	\begin{minipage}{0.49\linewidth}
		\centering
            \includegraphics[width=\linewidth]{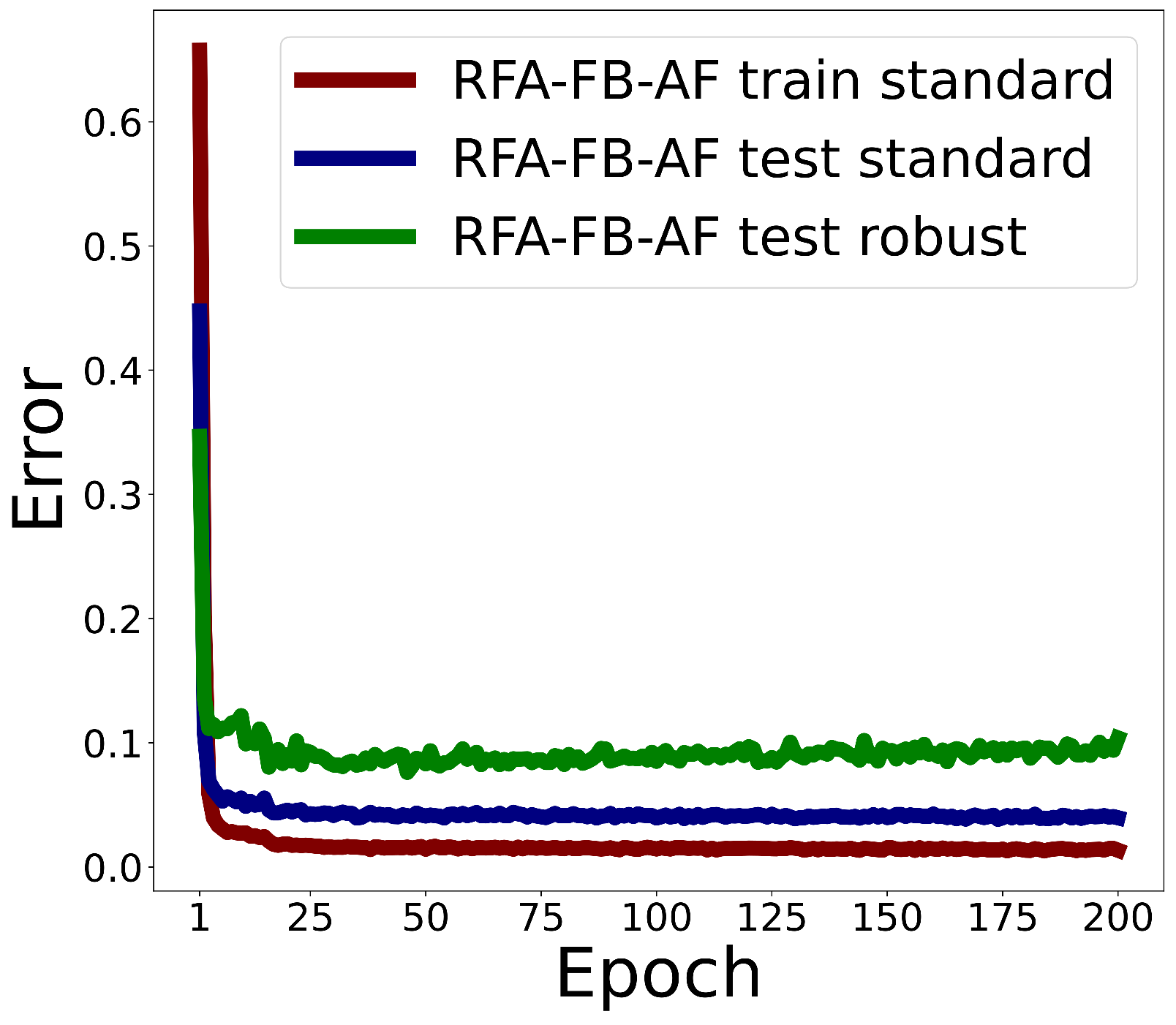}
		\textbf{(b)} RFA-FB-AF
	\end{minipage}
	\caption{Prediction errors by increasing epochs in AT.}
	\label{fig:RO}
\end{figure}

RFA is designed to overcome robust overfitting by introducing an additional learning path in the min-max game, where the inner loop is performed on the backbone to generate latent perturbations while the outer loop is optimized for fine-tuning the model by learning more robust features with RFA.

Figure \ref{fig:RO} (a) shows the RO effect with a popular AT method called TRADES \cite{Zhang2019} trained on CIFAR-10, where the AE perturbation for inner optimization is generated by PGD with $\epsilon=8/255$ in $k=10$ steps. 
It can be seen that the prediction error of both the standard training ('$\texttt{training standard}$') and standard test ('$\texttt{test standard}$') decreases with longer epochs of training on natural examples. 
Whereas the prediction error of test AE's ('$\texttt{test robst}$') starts to rise gradually after epoch 105, indicating RO in the AT process. 
As introduced in Section \ref{sec:RW}, RO often happens with AT methods and can seriously harm the test robustness generalization regardless of small and large models \cite{Rice2020}. Thus, remedies such as early stopping are suggested for AT \cite{Madry2018,Zhang2019}.
Note that in usual AT (e.g., TRADES) with full-model fine-tuning often significantly modifies pre-trained parameters and tends to cause larger error rates at the beginning of training in Figure \ref{fig:RO}(a).

Figure \ref{fig:RO}(b) shows the standard and robust test error rates of our method of RFA-based AT in feature space by fixing the backbone, namely RFA-FB-AF, with $g=3$ and $d=4$. The backbone model WRN-28-10 is also trained on CIFAR-10 dataset.  
It can be seen that RFA-FB-AF converges much faster in only about 25 epochs by updating the adapter parameters only. By continuing the training up to 50 epochs (i.e., twice the time after convergence), there is still \emph{no} RO phenomenon observed in Figure \ref{fig:RO}(b) where all test and training errors decrease with training iterations.

It is worth noting that RFA by itself is not a mere application of any existing adapters. The module is motivated and designed by viewing AT as a dynamic game \cite{Wang2023overfit}, and has to work in the min-max framework. The adapter helps to generate perturbations directly in the feature space and improves the inner-loop convergence quality by eliminating robust overfitting.  
As commented by the other reviewers, the proposed adapter-based method is novel and effective in the context of AT and offers a promising solution to address the issues of computational efficiency and robust overfitting.

\section{Experiments}

We evaluate the proposed RFA-based approach with several AT and other defense methods under various attacks on popular pre-trained models. 
In particular, we use WideResNet-28-10 (WRN) as the backbone model trained on CIFAR-10 and CIFAR-100 datasets \cite{krizhevsky_learning_2009}, Deit \cite{touvron_training_2021} of scales \texttt{Tiny}, \texttt{Small}, and \texttt{Base} trained on ImageNet (ILSVRC2012) \cite{Deng_ImageNet_2009} by its default setting with 1.28 million training image examples over 1000 classes. 
The optimizers are set to Adam with the learning rate initialized to 0.001. 
The hyper-parameters $\lambda_\mathtt{C_N}$, $\lambda_\mathtt{Tp}$, and $\lambda_\mathtt{B}$ for the loss functions in \eqref{eq:FB} and \eqref{eq:UB} are set to 0.4, 0.4, 0.6, respectively, based on empirical experiences. Unless otherwise specified, $g=3$ and $d=4$ for adding perturbation and RFA embedding. 
For fair comparison, \emph{all} AT methods use \textbf{PGD$_\infty$} to generate adversarial perturbations. The training batch size is 100 in each epoch and on V100 GPUs. 

Following \cite{croce2020reliable}, we evaluate the model's robust accuracy under a variety of diversified attacks sourced from the advertorch toolbox \footnote{\noindent\url{https://github.com/BorealisAI/advertorch}}. These include gradient-based PGD, optimization-based CW \cite{Carlini2017cw}, and adaptive combined method of AA \cite{croce2020reliable}. Specifically, we test with large attack strengths with $\epsilon=1$ for $l_2$ attacks, and $\epsilon=8/255$ for $l_\infty$ attacks. 
Both the initial constant and the confidence are set to 1.0 for CW$_2$. The number of steps for all attacks is set to 100.

\begin{table}[t!]
	\centering
	\caption{Model robustness under white-box attacks on CIFAR-10. \textbf{RFA}($\langle$backbone$\rangle$) operates in the RFA-FB mode.}
	\begin{tabular}{l|c|c|c|ccccc} 
		\bottomrule
		\multirow{2}{*}{Method} & \multirow{2}{*}{\begin{tabular}[c]{@{}c@{}}Clean\\Acc. (\%)\end{tabular}} & \multicolumn{5}{c}{Robust Accuracy (\%) under Attacks}                         \\ 
		\cline{3-7}
		&                                                                           & PGD$_\infty$  & PGD$_2$       & CW$_2$        & AA$_\infty$   & AA$_2$         \\ 
		\hline
		WRN                     & \textbf{95.5}                                                             & 0             & 0             & 0             & 0             & 0              \\
		AT-PGD \cite{Madry2018}                    & 85.7                                                                      & 56.0          & 32.5          & 7.5           & 48.4          & 21.9           \\
        
    	TRADES \cite{Zhang2019}                 & 84.8                                                                      & 60.9          & 35.5          & 8.2           & 45.3          & 22.6           \\
		FAT  \cite{zhang_attacks_2020}                   & 87.1                                                                      & 54.7          & 39.3          & 7.2           & 44.3          & 23.2           \\
		CDVAE \cite{Yang2021b}                  & 81.2                                                                      & 41.0          & 45.7          & 14.8          & 32.5          & 36.8           \\
		PAD \cite{zhou2023phase}                    & 83.6                                                                      & 39.7          & 32.9          & 10.5          & 25.0          & 17.4           \\
		SCORE \cite{pang_robustness_2022}                  & 88.1                                                                      & 70.6          & 44.25         & 7.7           & 60.9          & 31.3  \\
        DM-AT \cite{wang_better_2023}          & 92.9                   & 74.7     & 41.5   & 5.4    & 67.9    & 31.1    \\ 
        IKL-AT \cite{cui2024decoupled}
        & 92.2 				  & \textbf{75.3}  & 40.0   & 5.2  & 67.7  & 29.5  \\
        
		\hline
		\textbf{RFA}(WRN)       &  93.8                                                                    & 5.3           & 5.0           & 22.9          & 2.5           & 2.9            \\
		\textbf{RFA}(PGD)       & 84.1                                                                      & 53.6          & 32.0          & 53.8          & 51.6          & 27.1           \\
		\textbf{RFA}(FGSM)      & 84.4                                                                      & 53.5          & 37.9          & \textbf{66.6} & 50.8          & 37.1           \\
		\textbf{RFA}(DM)        & 92.1                                                                      & 73.7          & \textbf{51.4} & 33.4          & \textbf{72.2} & \textbf{40.1}  \\ 
		\hline

		\toprule
	\end{tabular}
	
	\label{tab:white}
\end{table}

\begin{table}[t!]
	\centering	
	\caption{Model robustness under white-box attacks on CIFAR-100.}
	\begin{tabular}{l|c|c|c|ccccc} 
		\bottomrule
		\multirow{2}{*}{Method} & \multirow{2}{*}{\begin{tabular}[c]{@{}c@{}}Clean\\Acc. (\%)\end{tabular}} & \multicolumn{5}{c}{Robust Accuracy (\%) under Attacks}                         \\ 
		\cline{3-7}
		&                                                                           & PGD$_\infty$  & PGD$_2$       & CW$_2$        & AA$_\infty$   & AA$_2$         \\ 
		\hline 
		WRN                                                 & \textbf{76.3}                  & 0        & 0      & 0      & 0       & 0       \\
		AT-PGD \cite{Madry2018} 
		                                                             & 58.4                   & 35.3     & 24.8   & 11.0     & 24.1    & 13.7    \\
		TRADES \cite{Zhang2019} 
		& 57.4                   & 39.3     & 26.5   & 11.6   & 26.4    & 14.3    \\
		FAT \cite{zhang_attacks_2020} 
		& 60.6                  & 35.9    & 29.8  & 11.3  & 26.9    & 18.8    \\
		CDVAE \cite{Yang2021b} 
		& 57.0                    & 18.0     & 22.7   & 10.9     & 12.9    & 14.8    \\
		PAD \cite{zhou2023phase}                    
		& 47.3                    & 26.6     & 25.8   & 10.2     & 13.3    & 12.9  \\
		SCORE \cite{pang_robustness_2022} 
		& 63.0                  & 45.6    & 29.8  & 12.5   & 31.3   & 16.9   \\
		
		DM-AT \cite{wang_better_2023} 
		                                                            & 73.0             & \textbf{51.7}     & 31.1  & 9.9    & 37.9   & 18.5   \\ 
		IKL-AT \cite{cui2024decoupled} 
                                                                    & 73.5 & 51.0    & 27.5 & 9.4  & 38.7 & 17.3  \\
		\hline
		\textbf{RFA} (WRN)                                                       & 73.1                  & 8.5      & 8.7   & 17.3  & 4.7    & 5.8    \\
		\textbf{RFA} (PGD)                                                                            & 59.9                   & 30.5     & 21.9   & 37.1   & 26.5    & 16.6    \\
		\textbf{RFA} (FGSM)                                                                         & 56.3                  & 30.1    & 25.0     & 31.6  & 26.6   & 22.3   \\
		
		\textbf{RFA} (DM)                                                                       & 71.2                 & 46.9     & \textbf{34.5}  & \textbf{37.2}  & \textbf{42.1}   & \textbf{25.5}   \\
		\toprule
	\end{tabular}
	
	\label{tab:white_cifar100}
\end{table}

\begin{table}[t!]
	\centering	
        \caption{Model robustness under gray-box attacks on CIFAR-10. All comparing methods use WRN as the backbone model.}
	\begin{tabular}{l|c|c|cccc} 
		\bottomrule
		\multirow{2}{*}{Method} & \multirow{2}{*}{\begin{tabular}[c]{@{}c@{}}Clean\\Acc. (\%)\end{tabular}} & \multicolumn{5}{c}{Robust Accuracy (\%) under Attacks}                         \\ 
		\cline{3-7}
		&                                                                           & PGD$_\infty$  & PGD$_2$       & CW$_2$        & AA$_\infty$   & AA$_2$         \\ 
		\hline
		AT-PGD \cite{Madry2018}        & 85.7                                                                      & 84.6          & 84.3          & \textbf{85.4} & 84.6          & 84.5           \\
		TRADES \cite{Zhang2019}        & 84.8                                                                      & 71.7          & 53.0          & 65.2          & 69.5          & 48.4           \\
		AdaAD \cite{Huang2023}                  & 83.3                                                                      & 67.2          & 49.8          & 37.6          & 64.2          & 46.3           \\
		HGD \cite{Liao2018}                    & 80.7                                                                      & 75.9          & 75.4          & 77.2          & 57.7          & -              \\
		APE-GAN  \cite{Jin2019}               & 90.9                                                                      & 59.3          & 65.2          & 65.3          & 56.0          & 59.8           \\
		CDVAE \cite{Yang2021b}                  & 86.8                                                                      & 77.1          & 78.0          & 78.3          & 74.4          & 75.8           \\
		PAD \cite{zhou2023phase}                    & 83.6                                                                      & 58.2          & 55.3          & 75.1          & 57.1          & 53.4           \\ 
		\hline
		\textbf{RFA}-FB-AE      & 94.3                                                                      & 97.9          & 94.6          & 65.8          & \textbf{93.8} & 92.7           \\
		\textbf{RFA}-FB-AF$_3$  & 93.8                                                                      & 93.1          & 92.2          & 46.5          & 92.8          & 91.3           \\ 
		\hline
		\textbf{RFA}-UB-AE      & \textbf{94.5}                                                             & \textbf{98.7} & \textbf{98.2} & 79.3          & 93.6          & \textbf{93.5}  \\
		\textbf{RFA}-UB-AF$_3$  & 93.7                                                                      & 95.3          & 96.7          & 60.0          & 93.0          & 90.5           \\
		\toprule
	\end{tabular}
	
	\label{tab:gray}
\end{table}

\begin{figure*}[t!]
	\centering
	\begin{minipage}{0.31\linewidth}
		\centering
		\includegraphics[width=.98\linewidth]{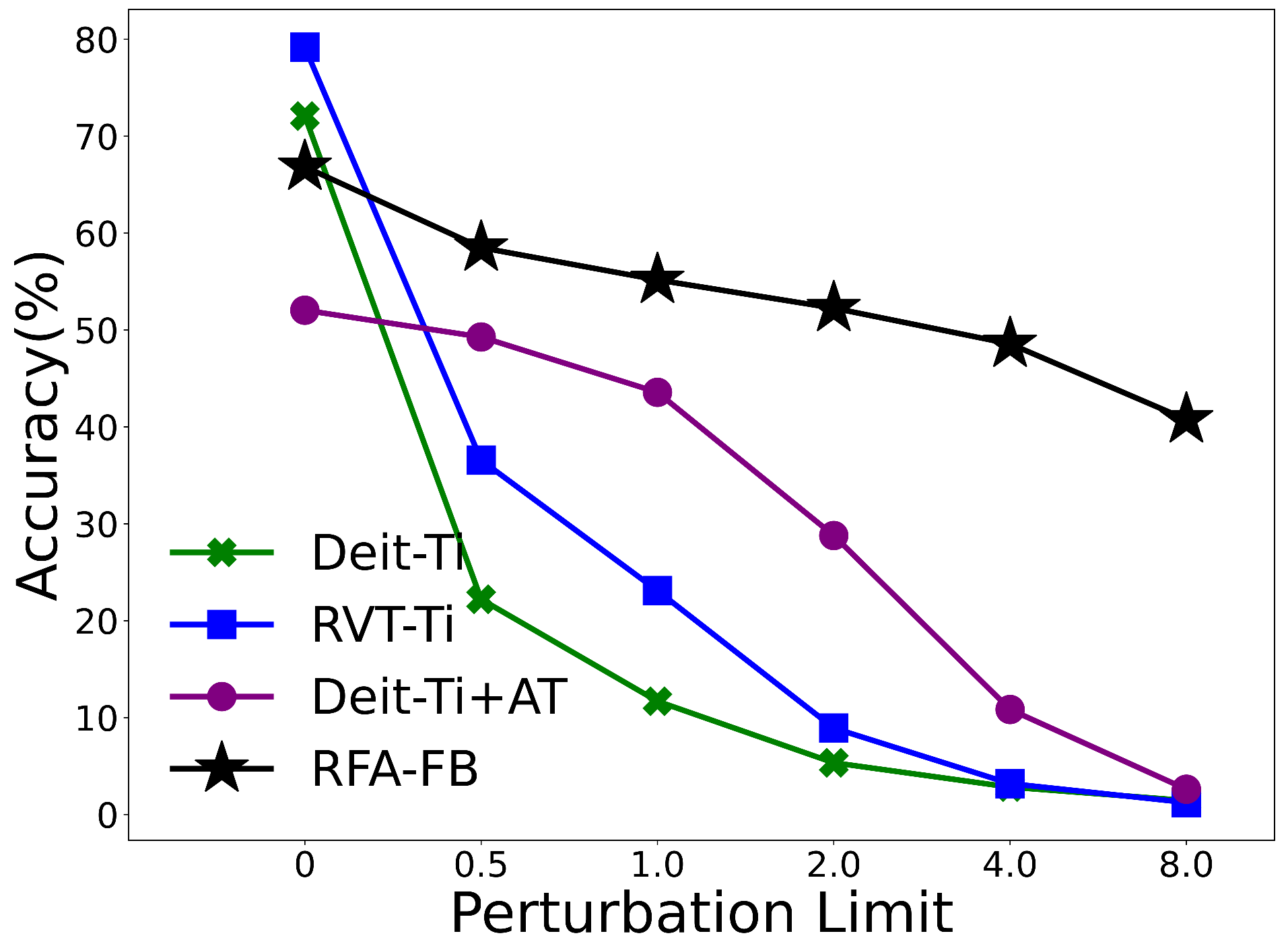}
		\textbf{(a)} $\mathtt{Tiny}$ 
	\end{minipage}
	\hfill
	\begin{minipage}{0.31\linewidth}
		\centering
		\includegraphics[width=.98\linewidth]{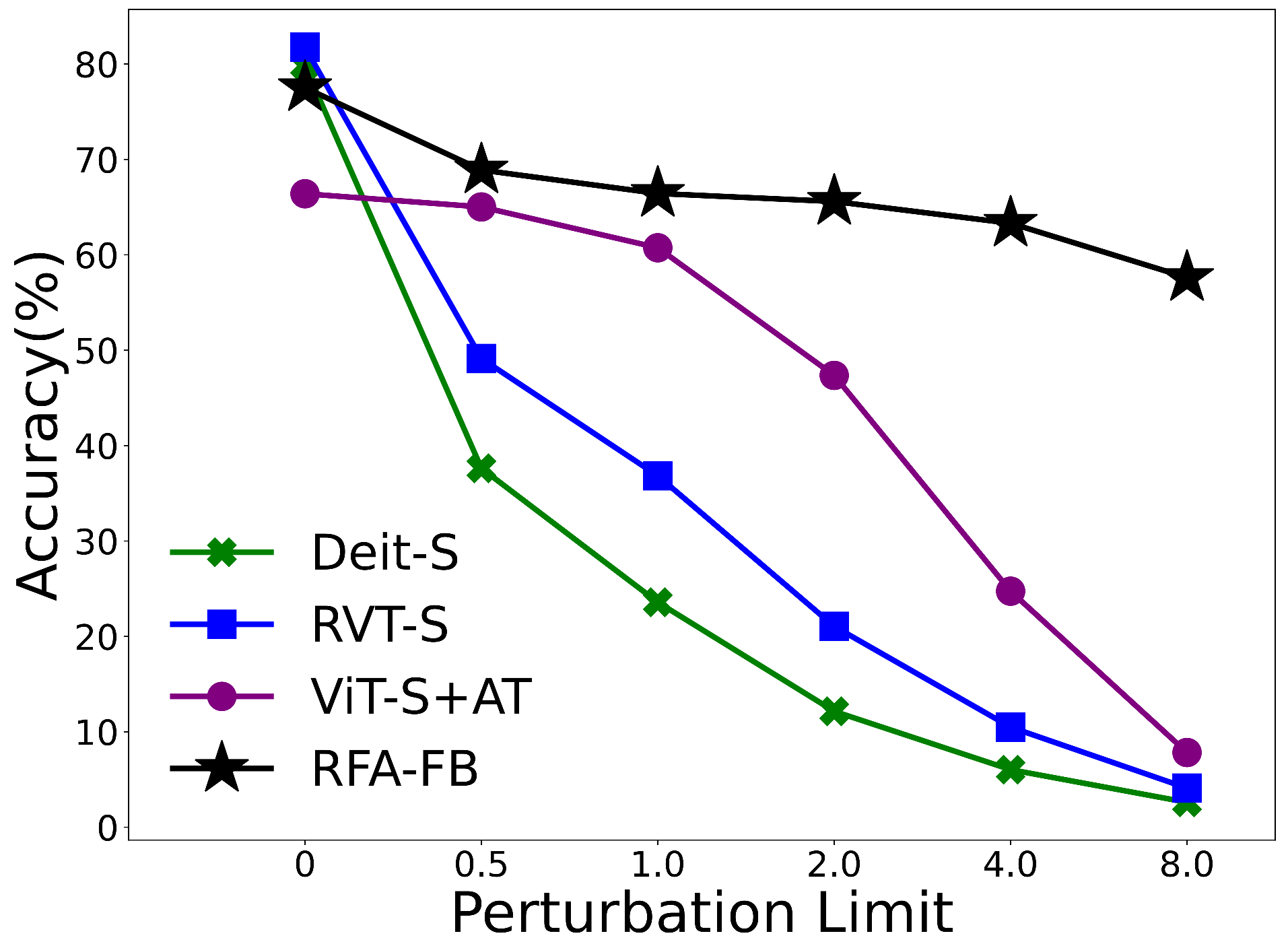}
		\textbf{(b)} $\mathtt{Small}$ 
	\end{minipage}
	\hfill
	\begin{minipage}{0.31\linewidth}
		\centering
		\includegraphics[width=.98\linewidth]{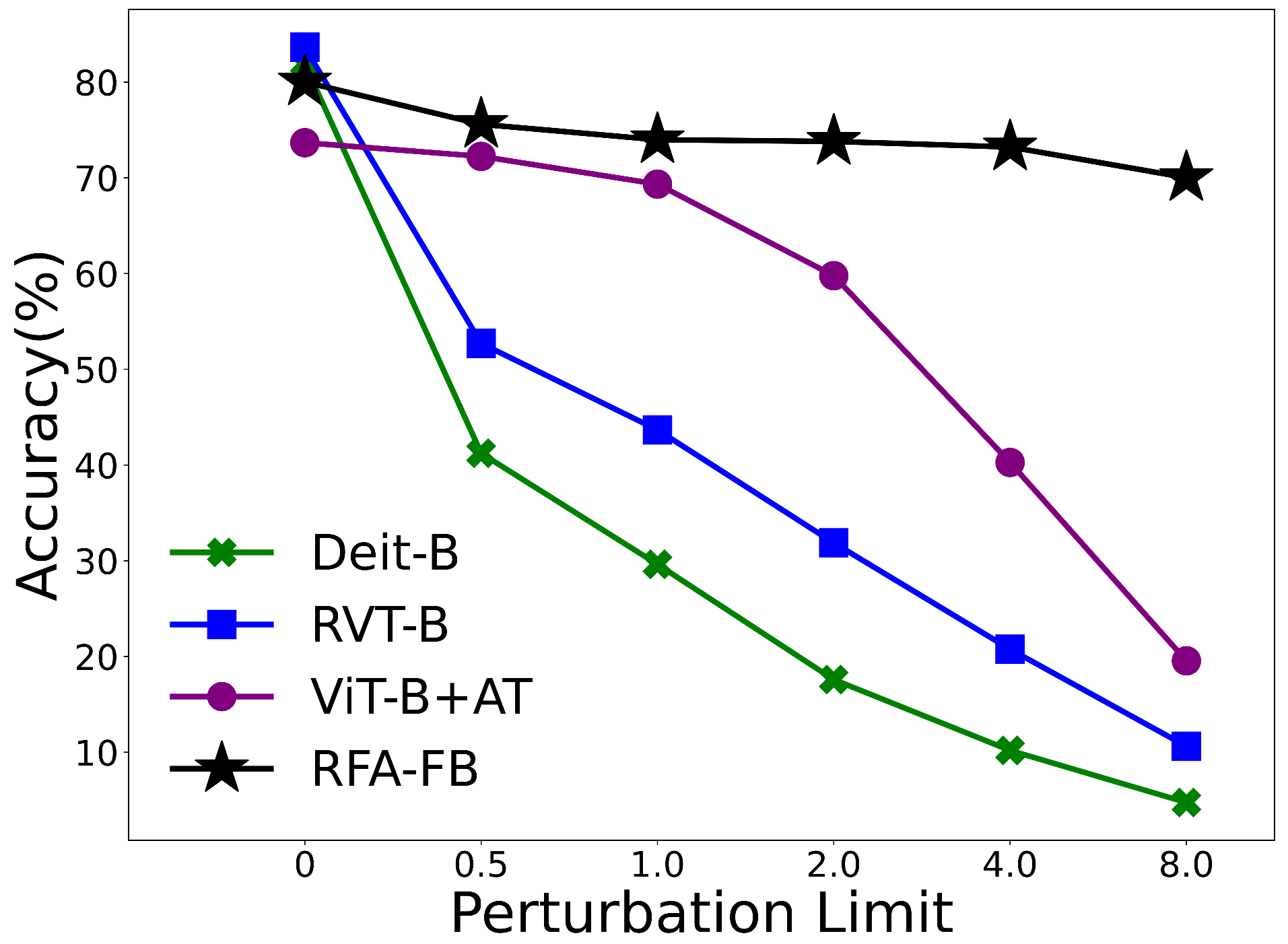}
		\textbf{(c)} $\mathtt{Base}$ 
	\end{minipage}
	\caption{Robust accuracy under PGD attacks with different perturbation limits on ImageNet of different scales. 
	}
	\label{fig:VIT_Robustness}
\end{figure*}

\subsection{White-Box Adversarial Robustness} 
\label{sec:robust}

We first test \emph{worst-case} scenarios using RFA-FB in the white-box setting.
In particular, we evaluate our method on several pretrained backbone models, denoted by \textbf{RFA} ($\langle$backbone$\rangle$) in Table \ref{tab:white}, respectively. The backbones can be either robust or non-robust models pre-trained on CIFAR-10. For example, CDVAE performs class disentanglement for input images and is incorporated with AT-PGD to boost the robust accuracy under white-box attacks \cite{Yang2021b}.

The model robustness is improved with RFA compared with that before incorporating RFA under most white-box attacks, especially with CW$_2$. 
In the worst case, the robust accuracy remains at about the same level comparable to that of the backbone model by itself. 
In Table \ref{tab:white}, it can be seen that the SOTA method of DM-AT \cite{wang_better_2023} and IKL-AT \cite{cui2024decoupled} excels on the PGD$_\infty$ attack used in AT but generalizes poorly to other unseen attacks, especially CW$_2$ of different mechanisms. 
Whereas RFA can boost the performance of robust models on unseen attacks by about 10-28\% at almost negligible cost of model accuracy on clean examples (i.e., clean accuracy), indicating better generalization abilities by our method.
In Table \ref{tab:white_cifar100}, RFA significantly enhances model robustness against a range of unseen attacks by 3-27\%, with only a minimal impact on clean accuracy, demonstrating better generalization capabilities that scale well to CIFAR-100.
 
\subsection{Gray-Box Adversarial Robustness} 
\label{sec:robust-gray-box}

We then evaluate AT methods in the gray-box setting which considers threats of using pre-trained models for downstream applications. It assumes that the attacker can access the backbone model but not the external adapter, e.g., by encrypting the RFA$_i$ module in our context.
We test both operation modes of RFA, i.e., RFA-FB and RFA-UB, respectively.  
The results are reported in Table \ref{tab:gray}.
Note that most existing AT methods modify the full pre-trained model parameters and cannot be used in the gray-box setting. 
For comparison, we consider a number of possible downstream defenses, e.g., by assuming the attacker does not know the student network in AdaAD \cite{Huang2023} or cannot access the pre-processing modules in \cite{Liao2018,Jin2019,zhou_towards_2021,Yang2021b,zhou2023phase}. 
For example, an amplitude-based preprocessing operation can be secured for phase-level AT in PAD \cite{zhou2023phase}. 
Table \ref{tab:gray} also includes transferred attacks generated on pre-trained models by baseline methods. 
 
The results show that both operation modes of RFA outperform the baselines by a large margin in gray-box settings. The reason may be two-fold. Firstly, the adapter-based approach better preserves the clean accuracy of pre-trained models (e.g., 95.5\% by WRN) even after robust training. The clean accuracy is reduced by only 1-2\% in our case compared with 5-15\% in all other defense methods by data pre-processing, adversarial distillation, and AT. This is consistent with the observations in Table \ref{tab:white} for better standard generalization by our method.
Secondly, our method is developed within the AT-PGD framework by introducing RFA to distill more robust features for better model generalization. 
It has inherent advantages in resisting PGD-like attacks including AA, as shown in Table \ref{tab:gray}. 
Comparing the two modes, RFA-UB-AE performs the best by updating the full model with perturbations generated in the input space.

We also compared our method with AdvXL \cite{Wang2024Revisiting} (pre-trained on ViT-H), Swin-B-AT \cite{liu2025comprehensive} and ViT-B-AT \cite{rebuffi2023revisiting} on ViT models in Table \ref{tab:robustVIT}. The experimental setting follows that in \cite{Wang2024Revisiting}, e.g.,  perturbation bounds of AutoAttack are set to $\epsilon_\infty=4/255, \epsilon_2=2.0$ to evaluate the robust accuracy on the ImageNet dataset. 
In general, larger models (with more parameters) exhibit stronger 
generalization capabilities. However, our method trained on a smaller Deit-B model still performs better under AutoAttack.

\begin{table}[t!]
	\setlength{\parindent}{1pt}
	\centering
	\caption{ViT model robustness on ImageNet.}
	\label{tab:robustVIT}
	\begin{tabular}{l|c|lll} 
		\bottomrule
		Methods         & Params(M) & Clean & AA$_\infty$ & AA$_2$  \\ 
		\hline
		ViT-B-AT \cite{rebuffi2023revisiting}    & \textbf{87}        & 76.7  & 53.5 & -     \\
		Swin-B-AT \cite{liu2025comprehensive}	& 88				 & 76.2  & 56.2 & 47.9 \\
		AdvXL(ViT-H) \cite{Wang2024Revisiting} & 304       & \textbf{83.9}  & 69.8 & 69.8  \\
		Ours(Deit-B)    & 94  & 80.0  & \textbf{70.8} & \textbf{71.5}  \\
		\toprule
	\end{tabular}
\end{table}

\subsection{Adversarial Robustness for AT at Scale}

We test the proposed RFA-FB method on the backbone model of Vision Transformer (ViT) at different scales. In particular, 
we use Deit \cite{touvron_training_2021} for the backbone model and deploy one transformer layer to replace VAE for feature distillation. 
The Tiny, Small, and Base scales of Deit are denoted by Deit-Ti, Deit-S, and Deit-B, respectively.
The experiments are conducted on ImageNet by its default setting with 1.28 million training image examples over 1000 classes.

Figure \ref{fig:VIT_Robustness} reports the adversarial robustness results under PGD$_2$ attacks with an increasing strength in comparison with the baseline ST model (Deit), standard AT \cite{Madry2018} (Deit + AT), and Robust Vision Transformer (RVT) \cite{mao_towards_2022}. 
In general, the proposed method of RFA-FB has the best test robustness for AT at all three scales of Tiny, Small, and Base. The robustness gain is significantly higher (e.g., $> 50\%$) than the comparing methods over the baseline, especially under stronger attacks with larger perturbation limits.

\subsection{Convergence Quality vs. Efficiency}
\begin{figure}[t!]
    \centering
    \includegraphics[width=\linewidth]{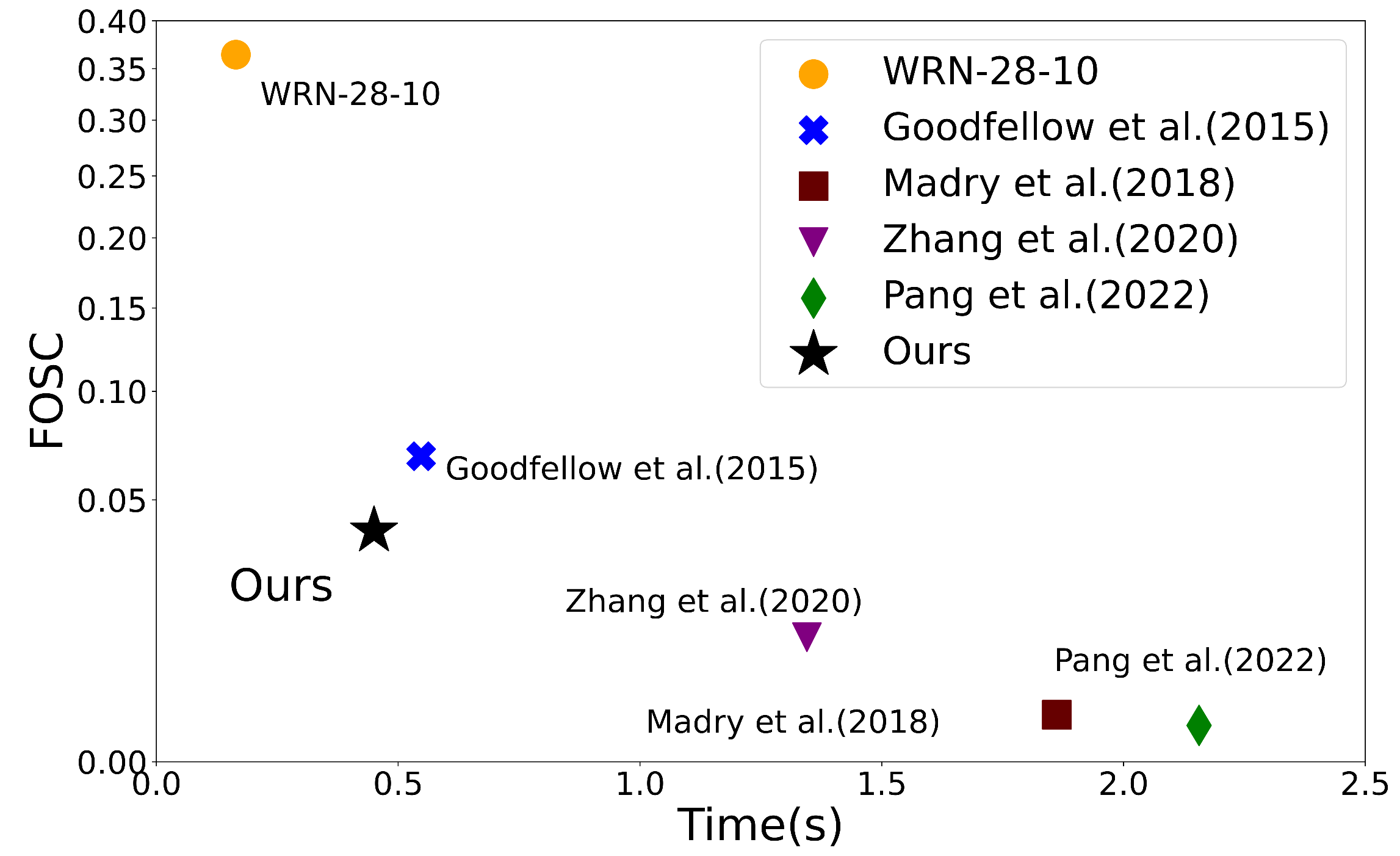}
    \caption{Inner-loop convergence quality vs. AT efficiency.}
    \label{fig:fosc_complexity}
\end{figure}
We report the training time per batch for the comparing methods in Table \ref{tab:white}. With the same batch size, RFA-FB-AF$_3$ is much faster by updating only the adapter parameters for fine-tuning regardless of the pre-trained backbone. The training time per batch is only 0.444 sec on CIFAR-10 and 0.4523 on CIFAR-100. 
The proposed adapter-based approach with a plug-in mode is more than five times faster than DM-AT \cite{wang_better_2023} and even 20\% faster than a single-step AT-FGSM method \cite{Jia2022a} in practice.

We can also evaluate the AT efficiency from another perspective. Following \cite{Wang2019FOSC}, we use the FOSC with PGD-10 to evaluate the convergence quality of inner optimization by different AT methods. 
It provides an indication of how well the AT performs in terms of inner-loop convergence quality. 
All comparing methods use WRN for the backbone and measure the computation efficiency by training time per batch in seconds.
Figure \ref{fig:fosc_complexity} plots the results. 

Both FOSC and training time favour smaller values. Therefore, method closer to the origin is more desirable in Figure \ref{fig:fosc_complexity}. Among all, \cite{pang_robustness_2022} achieves the minimum FOSC but pays the highest time cost;
\cite{zhang_attacks_2020} uses early stopped PGD for adversarial training, so it does a trade-off between time cost and FOSC.
Our method achieves the best trade-off between time cost and FOSC and is about four times faster than PGD on every training batch, and even faster than FGSM. The observation is consistent with the empirical results shown in Table \ref{tab:white} where our method significantly improves the training time efficiency by \emph{four} times faster compared with that of AT-PGD. 

Note that RFA-FB only needs to update the adapter parameters while fixing the backbone model. In Table \ref{tab:white}, for example, the training time of RFA is only 0.444 seconds per batch regardless of the backbone. With the same batch size, the training time per batch for AT-PGD and TRADES is 1.862 and 2.090 seconds, and that for more advanced defense methods such as FAT, CDVAE, PAD, and SCORE is 1.345, 2.017, 4.793, 2.169 seconds, respectively. In all cases, it can be seen that the proposed adapter-based approach is more than three times faster than the comparing method and is about 20\% faster than a single-step AT-FGSM method \cite{Jia2022a}.

\subsection{Computation Overhead}

To evaluate the computation overhead, we estimate the (additive) model complexity by the size of parameters (in Millions) and Floating Point Operations (FLOPs) required in testing 100 image examples. 
Table \ref{tab:cost_detection} shows that both measures are increased by only a slight fraction (i.e., less than $1/10$) of the inference cost by the backbone model, including a negligible computation overhead by including the RFA module for inference.

\subsection{Adversarial Detection}

The RFA is designed as a series adapter for AT. Once trained with a backbone model, the RFA module can work in a ``plug-in'' mode at inference time for adversarial detection by comparing the prediction results before and after the RFA embedding. 
We introduce a simple detector $\mathtt{D}(\cdot)$ consisting of a single MLP layer, i.e., $\mathtt{D}(\hat{y},\hat{y}_R)$.
Table \ref{tab:detect} provides experimental results that compare a number of adversarial detection methods including KD \cite{feinman2017detecting}, LID \cite{ma2018characterizing}, and MD \cite{lee2018simple}. Following \cite{Yang2021b}, we measure the true negative rate (TNR) at 95\% true positive rate and the area under an ROC curve (AUC). The RFA-based method outperforms the others in most cases.
We also test the detection accuracy against more than $100, 000$ AE's generated on ImageNet using PGD$_\infty$ with $\epsilon=8/255$ and PGD$_2$ with $\epsilon=1.0$. The detection accuracy of successful attacks is reported in Table \ref{tab:cost_detection}.

\begin{table}[t]
	\centering
	\caption{Computation overhead and adversarial detection on ViT. }
	\label{tab:cost_detection}
	\begin{tabular}
 {c|c|c|cc} 
		\bottomrule
		\multirow{2}{*}{Model} & \multirow{2}{*}{\begin{tabular}[c]{@{}c@{}}Param. \\Size (M)\end{tabular}} & \multirow{2}{*}{\begin{tabular}[c]{@{}c@{}}FLOPs (G)\end{tabular}} & \multicolumn{2}{c}{Detect. Acc.(\%)}  \\ 
		\cline{4-5}
		&                                                                      &                                                                     & $l_\infty$ & $l_2$                             \\ 
		\hline
		Deit-Ti (+RFA$_i$)         & 5.68 (+0.64)                                                          & 108 (+8.9)                                                           & 88.7    & 82.5                              \\
		Deit-S (+RFA$_i$)          & 21.9 (+2.16)                                                         & 425 (+35)                                                            & 92.9    & 83.8                              \\
		Deit-B (+RFA$_i$)          & 86.4 (+7.85)                                                         & 1687 (+140)                                                          & 94.6    & 86.3                              \\
		\toprule
	\end{tabular}
\end{table}

\begin{table}[t!]
 \setlength{\parindent}{1pt}
	\centering
	\caption{Adversarial detection accuracy (\%) on CIFAR-10.}
	\label{tab:detection}
	\begin{tabular}{l|c|c|c|c|c|c} 
		\bottomrule
		\multirow{2}{*}{Method} & \multicolumn{2}{c|}{FGSM} & \multicolumn{2}{c|}{PGD$_\infty$} & \multicolumn{2}{c}{PGD$_2$}  \\ 
		\cline{2-7}
		& TNR  & AUC                & TNR  & AUC                    & TNR  & AUC                  \\ 
		\hline
		KD                      & 42.4 & 85.7               & 73.1 & 94.6                   & 70.6 & 93.6                 \\
		LID                     & 69.1 & 93.6               & 71.5 & 93.2                   & 72.6 & 93.5                 \\
		MD                      & 94.9 & \textbf{98.7}               & 77.2 & 95.4                   & 76.7 & \textbf{95.3}                 \\
		RFA-FB                  & \textbf{97.6} & 96.9               & \textbf{95.6} & \textbf{96.1}       &            \textbf{92.4} & 94.4                 \\
		\toprule
	\end{tabular}
 
 \label{tab:detect}
\end{table}

\subsection{Ablation Study} 

We conduct an ablation study to evaluate the effects of  \eqref{eq:ce1}-\eqref{eq:tp}.
Table \ref{tab:ab} reports the results with a pre-trained WRN on CIFAR-10.
Similar to contrastive learning, both classifiers are necessary in the RFA training.
Without $\mathcal{L}_{\mathtt{C_N}}$, the negative samples used in $\mathcal{L}_{\mathtt{Tp}}$ are whatever features from $\mathtt{VAE_N}$ that may be entangled with positive samples and degrade the RFA performance.
Table \ref{tab:ab} shows that jointly training the dual-classifier with $\mathcal{L}_{\mathtt{Tp}}$ performs the best to generalize the adversarial robustness to unseen attacks, demonstrating the importance of including non-robust features in AT for robust generalization.

\begin{table}[t!]
	\centering
	\caption{Ablation study on the RFA learning.}
	\begin{tabular}{l|l|l|p{0.6cm}p{0.6cm}p{0.6cm}} 
		\bottomrule
		\multirow{2}{*}{Method} & \multirow{2}{*}{\begin{tabular}[c]{@{}l@{}}Clean\\Acc.\end{tabular}} & \multicolumn{4}{c}{Robust Accuracy (\%)}  \\ 
		\cline{3-6}
		&                                                                      & PGD$_\infty$  & AA$_\infty$ & AA$_2$ & DF$_\infty$                    \\ 
		\hline
		WRN							       & \textbf{95.5}  & 0.0 & 0.0   & 0.0   & 0.0 \\
		+$\mathcal{L}_{\mathtt{C_R}}$                  & 94.8  & 91.4  & 90.4  & 83.9  & 62.0 \\
		+$\mathcal{L}_{\mathtt{C_R}}$+$\mathcal{L}_{\mathtt{Tp}}$  & 89.8  & 89.1  & 88.3  & 83.6  & 59.4 \\ 
		+$\mathcal{L}_{\mathtt{C_R}}$+$\mathcal{L}_{\mathtt{C_N}}$+$\mathcal{L}_{\mathtt{Tp}}$                  & 93.8  & \textbf{93.1}  & \textbf{92.8}  & \textbf{91.3}  & \textbf{70.8} \\ 
		\toprule
	\end{tabular}

        \label{tab:ab}
\end{table}

\section{CONCLUSION}

We proposed a new adapter-based method to perform efficient AT directly in the feature space. The RFA module is designed to overcome robust overfitting by generating and incorporating perturbations directly in the feature space. This results in significantly better robust generalization against unseen attacks with better convergence quality and negligible computation overhead.
It can work with other defense methods to imbue adversarial robustness at scale over different model architectures including CNN and ViT. The RFA can also be easily used to perform adversarial detection by ``plug-in'' for testing at inference time.


\section{Technical Appendix}
\label{sec:appendix}

\subsection{A1. Proof of Proposition 1}
\label{sec:ap1}

Let $f_l(\cdot)$ be the transformation function in the hidden layer $l$, $l=0, 1, ..., L$, which usually consists of a linear operation and a non-linear activation. 
Specifically, we have $f_0(x) = x$ being the input layer and $f_L$ being the softmax layer for model prediction.
Without loss of generality, the backbone model can be expressed as $B(x) = f_L(...f_1(f_0(x)))$, and its fraction $B_l(x) =  f_l(...f_1(f_0(x)))$. 

We first discuss the accumulated effect on the output layer of $f_L(\cdot)$ by adding perturbation $\delta_0$ to the input layer $f_0(x)$ .  
Let $z_0 = x + \delta_0$. 
If all values in the Hessian of $f_1(z_0)$ are finite, i.e., $|\partial^2 f_1(z_0)_i/\partial z_j \partial z_k| < \infty, \forall i, j, k$, then the perturbation effect can be approximated with the first-order Taylor's expansion on $f_1(z_0)=f_1(f_0(x)+\delta_0)$ at the input $x$, i.e., 
$f_1(z_0)\approx f_1(x) + \nabla_x f_1(x) \cdot \delta_0$. Let $z_1 = f_1(x)$ and $\delta_1 = \nabla_x f_1(x) \cdot \delta_0$. Repeating the process for the subsequent layers recursively and assuming that all Hessians of the form $\nabla^2 f_l(x)|_{f_m(x)}, \forall m<l$ are finite, we have the approximation of the accumulated effect of $\delta_0$ at the output layer as $B(x+\delta_0) \approx B(x) + \nabla_x B(x) \cdot \delta_0$. 
For simplicity, we drop the notation $y$ in the cross-entropy loss function $\mathcal{L}(\cdot)$ hereafter.
The resulting variation of the CE loss caused by $\delta_0$ is therefore 
\cite{Simon-Gabriel2019}
\begin{equation}
\label{eq:x}
	\begin{split}
		\Delta\mathcal{L}_0 &= \max_{\|\delta_0\|_\infty\leq \epsilon} | \mathcal{L} \left(B[x+\delta_0]\right) -  \mathcal{L} \left(B[x]\right) | \\
		&\approx  \max_{\|\delta_0\|_\infty\leq \epsilon} |\langle  \delta_0, \nabla_x  \mathcal{L} \left(B[x]\right) \rangle | \\
		&= \epsilon \|  \nabla_x  \mathcal{L} \left(B[x]\right) \|_1
	\end{split}
\end{equation}
where $\langle\cdot,\cdot\rangle$ represents the inner product operation. Note that $\|\cdot\|_1$ is the dual norm of $\|\cdot\|_\infty$. 

Similarly, by imposing perturbations $\delta_g$ on the intermediate feature $z_g$ at the $g$-th sub-layer, we study its effect through all subsequent layers to the output, denoted by $B_{g+}(z_g) = f_L(...f_g(z_g))$. 
Consider all inputs to $B_{g+}$ as vectors. Therefore, for $g>0$, we have
\begin{equation}
\label{eq:g}
	\begin{split}
		\Delta\mathcal{L}_g &= \max_{\|\delta_g\|_\infty} 
  | \mathcal{L} \left(B_{g+}[z_g+\delta_g]\right) - \mathcal{L} \left(B_{g+}[z_g]\right) | \\
		&\approx  \max_{\|\delta_g\|_\infty}
  |\langle  \delta_g, \mathbf{J}(B_{g+}[z_g]) \rangle | \\
&=\max \|\delta_g\|_\infty \cdot \|\mathbf{J}(B_{g+}[z_g])\|_1
	\end{split}
\end{equation}
where $\mathbf{J}(B_{g+}[z_g])$ is the $g$-th layer Jacobian of the network function $B_{g+}[\cdot]$.

On the other hand, suppose that perturbations $\delta_h$, 
for $h<g$, is added to the $h$-th sub-layer instead.
Assuming Hessian with finite elements, 
we can estimate its effect on the $g$-th layer $z_g = B_{hg}(z_h+\delta_h)$, where $B_{hg}(\cdot) = f_g(...f_h(\cdot))$, by taking the first-ordered expansion on $z_g$. Thus, the approximated perturbation is denoted by
\begin{equation}
    \tilde{\delta}_g = \langle \delta_h, \nabla_{z_h} z_g\rangle 
     = \left\langle \delta_h, \mathbf{J}(B_{hg}[z_h])\right\rangle 
\end{equation}
Similar to \eqref{eq:g}, we have
\begin{equation}
\label{eq:h}
	\begin{split}
		\Delta\mathcal{L}_h &= \max_{\|\delta_h\|_\infty} | \mathcal{L} \left(B_{h+}[z_h+\delta_h]\right) - \mathcal{L}\left(B_{h+}[z_h]\right) | \\
		&\approx \max_{\|\tilde{\delta}_g\|_\infty} | \mathcal{L} (B_{g+}[z_g+\tilde{\delta}_g]) - \mathcal{L}\left(B_{g+}[z_g]\right) | \\
		&= 
         \max \|\tilde{\delta}_g\|_\infty \cdot \|\mathbf{J}(B_{g+}[z_g])\|_1 
	\end{split}
\end{equation}  

Now we compare the budgets of $\|\tilde{\delta}_g\|_\infty$ and $\|\delta_g\|_\infty$. In our setting, $\|\delta_g\|_\infty\leq k\eta\cdot\|z_g\|_1$ and $\|\delta_h\|_\infty\leq k\eta\cdot\|z_h\|_1$ have the same hyper-parameter values of $k$ and $\eta$.  
According to the H\"{o}lder's inequality, we have
\begin{equation}
\label{eq:est}
\begin{split}
    \| \tilde{\delta}_g \|_\infty &= \max \|\left\langle \delta_h, \mathbf{J}(B_{hg}[z_h]) \right\rangle \|_1 \\
   &\leq \max \|\delta_h\|_\infty \cdot \|\mathbf{J}(B_{hg}[z_h])\|_1 \\
   &= k\eta\cdot\|z_h\|_1\cdot \|\mathbf{J}(B_{hg}[z_h])\|_1 
\end{split}
\end{equation} 

Assume Lipschitz continuity of the backbone network function $f(\cdot)$ with two different inputs $x$ and $y$, i.e.,  
\begin{equation}
    \| f(x) - f(y) \|_1 \leq c \cdot \| x - y\|_1 
    \label{eq:lip}
\end{equation}
which implies that the network output does not fluctuate much in response to small changes in the input. In this case, the $l_1$-norm of the Jacobian $\mathbf{J}(f(x))$ is usually bounded and can be controlled by the Lipschitz constant $c$. Accordingly, the following inequality holds for such an $f(x)$, which is of the form \cite{tsuzuku2018lipschitz}
\begin{equation}
    \|f(x)\|_1 \leq \|x\|_1\cdot \|\mathbf{J}(f(x))\|_1
    \label{eq:ineq}
\end{equation}
Therefore, we have $\|z_g\|_1\leq \|z_h\|_1\cdot \|\mathbf{J}(B_{hg}[z_h])\|_1$ and thus $\max \| \delta_g \|_\infty \leq \max \| \tilde{\delta}_g \|_\infty$. From \eqref{eq:g} and \eqref{eq:h}, it can be seen that $\Delta\mathcal{L}_h > \Delta\mathcal{L}_g$ with $h<g$ and the same hyper-parameter value of $k\eta$ for the setting of perturbation budgets, under the assumptions of Hessian with finite elements and Lipschitz continuity for the backbone network layers. This completes the proof.

\subsection{A2. Robust and Non-Robust Features by RFA}
\label{ap:GANForAFs}

\begin{figure}[t!]
	\centering 
        \includegraphics[width=\linewidth]{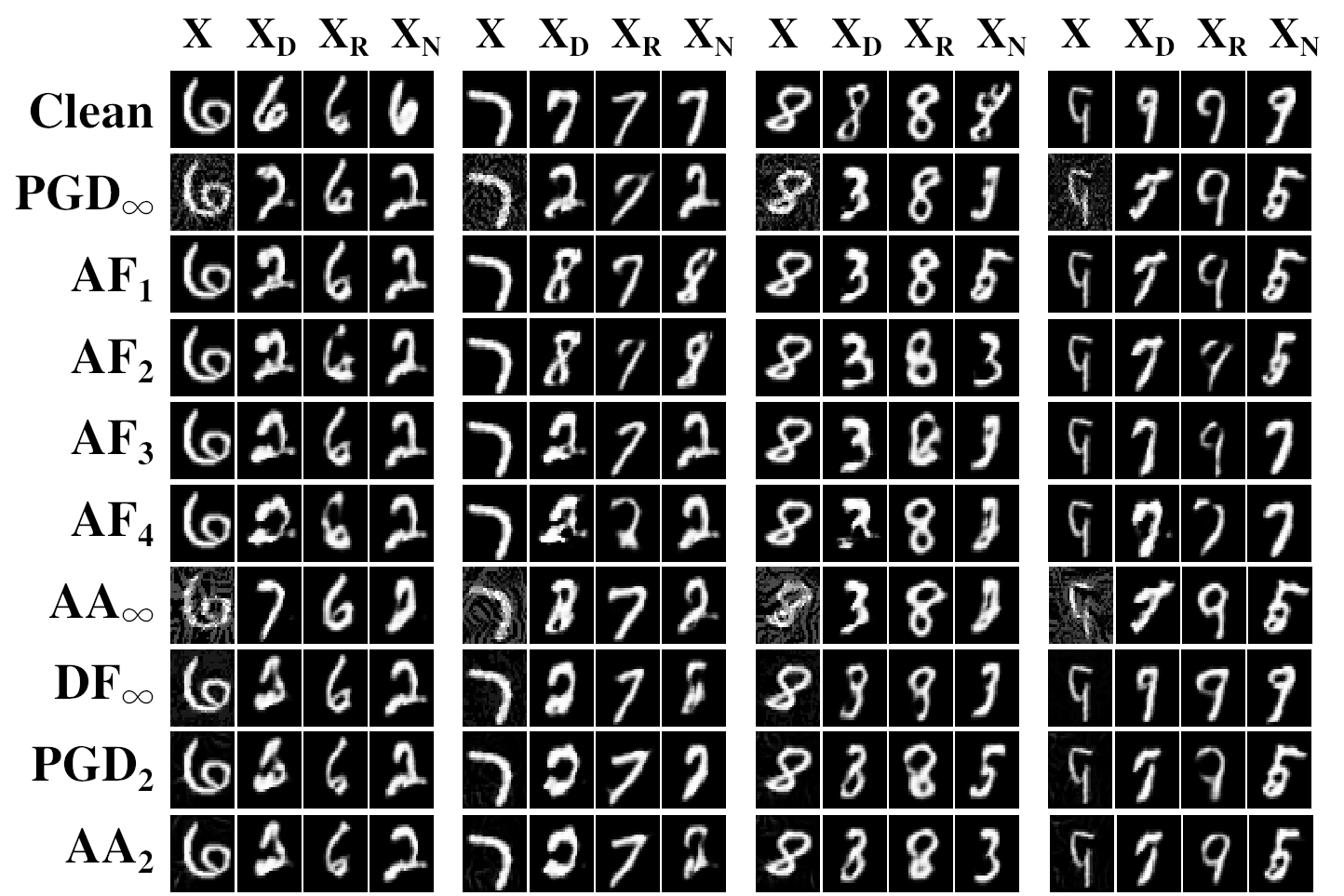}
	\caption{Image Reconstructions from the RFA-learned robust and non-robust features by LSGAN.}
	\label{fig:feature_gan}
\end{figure}

\begin{table}[!t]
	\centering
	\caption{Correlation of the predictive outputs of $\mathtt{C_D}$, $\mathtt{C_R}$, and $\mathtt{C_N}$ on normal examples ($+$) and adversarial examples ($-$).}
	\begin{tabular}{c|p{0.8cm}|p{0.5cm}|p{0.5cm}p{0.5cm}p{0.5cm}p{0.6cm}} 
		\bottomrule
		\multirow{2}{*}{Correlation}                       & \multicolumn{2}{c|}{Inner Attacks} & \multicolumn{4}{c}{Unseen Attacks}                                      \\ 
		\cline{2-7}
		& PGD$_\infty$ & AF$_1$ & PGD$_2$ & AA$_\infty$ & AA$_2$ & DF$_\infty$  \\ 
		\hline
		$\cos(\hat{y}^-_\mathtt{N}, \hat{y}^-_\mathtt{D})$ & 0.614 & 0.823 & 0.852  & 0.571 & 0.843 & 0.712                      \\
		$\cos(\hat{y}^-_\mathtt{R}, \hat{y}^+_\mathtt{D})$ & 0.948 & 0.920 & 0.825  & 0.912 & 0.855 & 0.877                      \\
		$\cos(\hat{y}^-_\mathtt{R}, \hat{y}^-_\mathtt{N})$ & 0.009 & 0.061 & 0.181  & 0.034 & 0.126 & 0.362                      \\
		\toprule
	\end{tabular}
	
	\label{tab:correlation}
\end{table}

We first perform image reconstruction to visualize semantic changes of the $d$-th sub-layer feature under attacks. 
In this experiment, we use ResNet18 as the backbone and perform AT with RFA on MNIST \cite{lecun1998gradient}. The sub-layer for RFA embedding is $d=5$. 
Following notations in the paper, let $z^{+}_d$ be the $d$-th layer feature of input image $X$. Its adversarial counterpart $z^{-}_d$
is decomposed by RFA into the robust and non-robust components, denoted by $z^{-}_\mathtt{R}$ and $z^{-}_\mathtt{N}$, respectively. 
We use LSGAN \cite{mao_least_2017} to reconstruct images based on the intermediate features of $z_d$, $z_\mathtt{R}$ and $z_\mathtt{N}$, respectively. 
The results are shown in Figure \ref{fig:feature_gan}.  

Without attacks (Clean), the reconstructed images corresponding to all the intermediate features, i.e., $z^{+}_d$, $z^{+}_\mathtt{R}$ and $z^{+}_\mathtt{N}$, have correct semantics as the original input $X$.  
Under attacks, the semantic information of $X_\mathtt{D}$ and $X_\mathtt{N}$ starts to alter for $z^{-}_d$ and $z^{-}_\mathtt{N}$ while $X_\mathtt{R}$ remains the same for $z^{-}_\mathtt{R}$. The experimental results in general verify that the RFA is able to distill class-representative robust features into $z_d$.

In particular, rows with AF$_g$ present visualization results by adding perturbations to the $g$-th layer.  
In most cases, $X_\mathtt{N}$ holds similar semantic information to $X_\mathtt{D}$. This supports the novel design of RFA that exploits $z_\mathtt{N}$ to estimate the distribution of vulnerable features such that $z_\mathtt{R}$ can be \emph{pushed away} from attacks as far as possible.

Table \ref{tab:correlation} shows quantitative results to further demonstrate the effect.
In particular, the predictive output of $\mathtt{C_N}$ on the non-robust features, i.e., $\hat{y}^{-}_\mathtt{N}:=\mathtt{C_N}(B_{d+}[z^{-}_\mathtt{N}])$, is highly correlated with that of $\mathtt{C_D}$ on AE, i.e., $\hat{y}^{-}_\mathtt{D}:=\mathtt{C_D}(B_{d+}[z^{-}_d])$, in terms of their $\cos$ values. 
With the same learning functions, this indicates that the adversarial feature perturbation affecting $z_d$ with the class-representative information is primarily extracted into the non-robust features in $z^{-}_\mathtt{N}$. 
On the other hand, the predictions of $\mathtt{C_R}$ on the robust features, i.e., $\hat{y}^{-}_\mathtt{R}:=\mathtt{C_R}(B_{d+}[z^{-}_\mathtt{R}])$, and those of $\mathtt{C_D}$ on normal examples, i.e., $\hat{y}^{+}_\mathtt{D}:=\mathtt{C_D}(B_{d+}[z^{+}_d])$, exhibit high correlations. With a small $\cos(\hat{y}^{-}_\mathtt{R}, \hat{y}^{-}_\mathtt{N})$, it can be inferred that $z^{-}_\mathtt{R}$ is pushed away from $z^{-}_\mathtt{N}$ by the RFA learning in terms of their class-representative information.
The RFA-learned features also generalizes well to unseen attacks, such as AA with $l2$ and $l_\infty$ norms and DF$_\infty$ as shown in Figure \ref{fig:feature_gan} and Table \ref{tab:correlation}. 

Interestingly, under adversarial attacks, the $z^-_\mathtt{N}$ obtained by decoupling $z^-_\mathtt{d}$ through RFA exhibits a slight discrepancy. Table \ref{tab:correlation} shows that $\cos(\hat{y}^-_\mathtt{N}, \hat{y}^-_\mathtt{D})$ under PGD$\infty$ and AA$\infty$ attacks displays a slight discrepancy, which is also consistent with the results shown in Figure \ref{fig:feature_gan}. However, this phenomenon aligns with our goal of designing RFA to extract more diverse non-robust features.

\footnotesize{\bibliography{m9787}}

\end{document}